\documentclass[10pt,twocolumn,letterpaper]{article}

\usepackage[pagenumbers]{cvpr} 

\usepackage[dvipsnames]{xcolor}

\definecolor{cvprblue}{rgb}{0.21,0.49,0.74}
\usepackage[pagebackref,breaklinks,colorlinks,citecolor=cvprblue]{hyperref}
\usepackage{multirow}
\usepackage{pifont}
\usepackage{makecell}
\usepackage{xcolor}
\usepackage{algorithm}
\usepackage{algorithmic}
\definecolor{darkgreen}{rgb}{0.17,0.56,0.36}

\def\paperID{1161} 
\def\confName{CVPR}
\def\confYear{2024}

\newcommand{\mono}{Mono-M}
\newcommand{\avg}{Dual-M}
\newcommand{\proposed}{S2DHand}
\newcommand{\upscore}[1]{\footnotesize{\mydarkgreen{$\blacktriangledown$ #1\%}}}

\newcommand{\mydarkgreen}[1]{\textcolor[rgb]{0.17,0.56,0.36}{ #1}}

\title{Single-to-Dual-View Adaptation for Egocentric 3D Hand Pose Estimation}

\author{{Ruicong Liu \qquad Takehiko Ohkawa \qquad Mingfang Zhang \qquad Yoichi Sato}\\
The University of Tokyo, Tokyo, Japan\\
{\tt\small \{lruicong, ohkawa-t, mfzhang, ysato\}@iis.u-tokyo.ac.jp}
}

\begin{document}
\maketitle
\begin{abstract}
The pursuit of accurate 3D hand pose estimation stands as a keystone for understanding human activity in the realm of egocentric vision.
The majority of existing estimation methods still rely on single-view images as input, leading to potential limitations, e.g., limited field-of-view and ambiguity in depth.
To address these problems, adding another camera to better capture the shape of hands is a practical direction.
However, existing multi-view hand pose estimation methods suffer from two main drawbacks: 
1) Requiring multi-view annotations for training, which are expensive. 
2) During testing, the model becomes inapplicable if camera parameters/layout are not the same as those used in training.
In this paper, we propose a novel Single-to-Dual-view adaptation (\proposed) solution that adapts a pre-trained single-view estimator to dual views.
Compared with existing multi-view training methods, 1) our adaptation process is unsupervised, eliminating the need for multi-view annotation.
2) Moreover, our method can handle arbitrary dual-view pairs with unknown camera parameters, making the model applicable to diverse camera settings.
Specifically, \proposed~is built on certain stereo constraints, including pair-wise cross-view consensus and invariance of transformation between both views.
These two stereo constraints are used in a complementary manner to generate pseudo-labels, allowing reliable adaptation.
Evaluation results reveal that \proposed~achieves significant improvements on arbitrary camera pairs under both in-dataset and cross-dataset settings, and outperforms existing adaptation methods with leading performance. Project page: \url{https://github.com/ut-vision/S2DHand}.
\end{abstract}
\vspace{-1em}    
\section{Introduction}
\label{sec:intro}


Delving into the realm of egocentric vision (first-person view), the pursuit of refining 3D hand pose estimation stands as a keystone for understanding human activity. 
This quest not only forges new paths in human-computer interaction \cite{H:pavlovic1997visual,H:rautaray2015vision,H:von2001bare}, but also empowers imitation learning \cite{H:garcia2020physics,H:handa2020dexpilot,H:vogt2007prefrontal}. 
Moreover, it enhances the immersive experience in augmented/virtual reality (AR/VR) to new heights \cite{H:holl2018efficient,H:piumsomboon2013user}. 
Recently, with the advancements of AR/VR headsets, egocentric data has become increasingly prevalent \cite{E:damen2018scaling,E:grauman2022ego4d}, leading to an increasing demand for estimating 3D hand poses from egocentric viewpoints.

To achieve better 3D hand pose estimation performance, recent years have witnessed many networks with various structures \cite{H:wen2023hierarchical,H:zhou2020monocular,H:fu2023deformer}.
However, the majority of existing hand pose estimation methods are still under a single-view setting, which is convenient but leads to potential limitations, \eg, limited field-of-view and ambiguity in depth.
To address these problems, a potential solution is to add another camera to expand the field-of-view and reduce depth ambiguity by capturing the hand shape from an additional view angle.
Furthermore, the use of multiple cameras also aligns with industry trends, as demonstrated by the latest AR/VR headsets such as the Apple Vision Pro and Meta Quest, which feature multiple egocentric cameras.
Overall, an unavoidable trend towards multi-view settings in hand pose estimation is emerging, driven by its technological advantages and the direction of industrial development.

\begin{figure}[t]
	\centering
	\includegraphics[width=\linewidth]{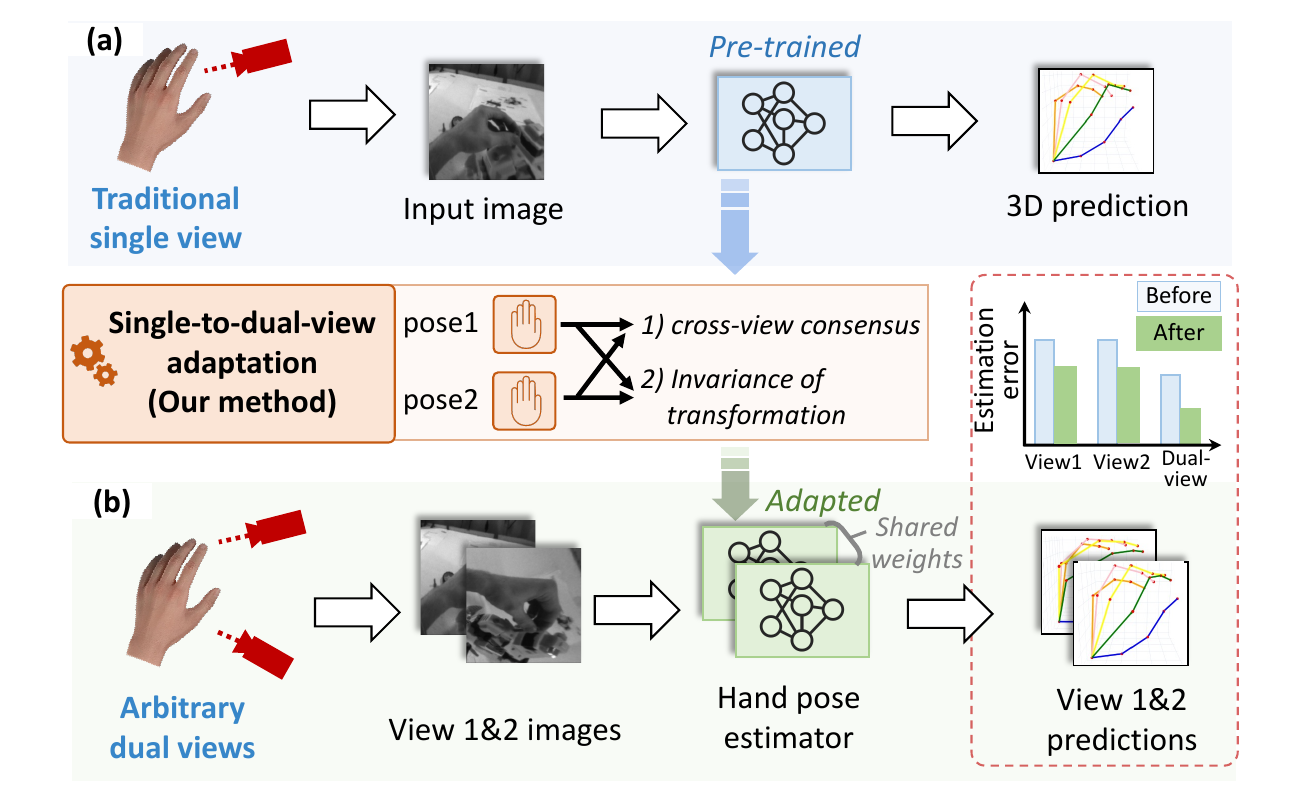}
	\caption{From (a) to (b), our single-to-dual-view adaptation method adapts a traditional single-view hand pose estimator to arbitrary dual views. The adapted model becomes more accurate under the dual-view setting. (a) Traditional single-view hand pose estimation. (b) Inference process of the adapted model under a dual-view setting.}
	\label{fig:teaser}
	\vspace{-6mm}
\end{figure}

\begin{table*}[htbp]
\begin{center}
\caption{Differences among traditional single-view methods, multi-view training methods, and our proposed single-to-dual-view adaptation method. Green indicates lower training requirements, enhanced testing results, and reduced camera parameter requirements, respectively.}
\label{tab:diff}
\vspace{-3mm}
\setlength{\tabcolsep}{1mm}
{   \small
    \resizebox{\linewidth}{!}{
    \begin{tabular}{llll}
        \bottomrule[1.2pt]
        \specialrule{0em}{1pt}{1pt}
        Methods & Pre-training dataset required & Test  & Camera parameters\\
        \hline
        \specialrule{0em}{1pt}{1pt}
        Traditional single-view methods \cite{H:wen2023hierarchical,H:zhou2020monocular,H:fu2023deformer} & \textcolor{darkgreen}{Single-view} (\textcolor{darkgreen}{common single-camera} setting) & \textcolor{red}{Single-view} & \textcolor{darkgreen}{Not required} \\
        Multi-view training methods \cite{H:chen2021mvhm,H:khaleghi2022multi,H:han2022umetrack} & \textcolor{red}{Dual-view} (need \textcolor{red}{multi-camera} setting) & \textcolor{darkgreen}{Dual-view} \& \textcolor{red}{same} camera poses with training & \textcolor{red}{Required} \& \textcolor{red}{same} with training \\
        Single-to-dual-view adaptation (Ours) & \textcolor{darkgreen}{Single-view} (\textcolor{darkgreen}{common single-camera} setting) & \textcolor{darkgreen}{Dual-view} \& \textcolor{darkgreen}{arbitrary} camera poses & \textcolor{darkgreen}{Not required}\\
        \bottomrule[1.2pt]
\end{tabular}  }
}
\end{center}
\vspace{-7mm}
\end{table*}

Currently, several existing studies \cite{H:chen2021mvhm,H:khaleghi2022multi,H:han2022umetrack} have paid attention to hand pose estimation under multi-view settings.
These methods typically process input images from multiple views simultaneously, utilizing a feature fusion module to arrive at a final prediction \cite{N:liu2023multi,N:xu2021multi}.
However, all these methods have two significant drawbacks that limit their applicability. 
1) The training, especially for the feature fusion module, necessitates multi-view labels, which are costly to annotate.
2) During testing, the same camera parameters as in training must be used.
An estimator trained under a specific multi-camera setup becomes inapplicable if there are any changes to the camera layout or parameters.

Unlike existing multi-view training methods, we propose a new solution that adapts an estimator from single-view to dual-view without needing multi-view labels or camera parameters. 
As shown in \cref{fig:teaser}, given a pre-trained estimator, our method adapts it to an arbitrary dual-view setting (from (a) to (b)), where two cameras are placed in any layout without knowing their parameters.
Here, all we need is a pre-trained estimator and a sufficient number of unlabeled dual-view inputs from the two cameras. 
As compared in \cref{tab:diff} (row 2-3), in contrast to multi-view training, our method only needs common and cheaper single-view data for training. 
During testing, unlike existing methods, our method is compatible with arbitrary dual-view pairs, making the model applicable to flexible and changeable camera settings.
Specifically, when camera settings change, it is easy and swift to repeat our method’s adaptation process to re-adapt the pre-trained estimator to work well with new camera parameters.
For camera parameters, existing methods not only need them for training, but also require them the same with testing.
Conversely, our method is clearly more practical since no camera parameters are required.

Building on these advancements, we present a novel unsupervised Single-to-Dual-view adaptation framework (\proposed) for egocentric 3D hand pose estimation.
It uses certain stereo constraints for adaptation, including cross-view consensus (pair-wise) and invariance of transformation between both camera coordinate systems (to all input pairs). 
These two stereo constraints are used in a complementary manner to refine the accuracy of pseudo-labels, allowing the model to better fit to the dual views.
Specifically, the cross-view consensus is leveraged through an attention-based merging module, and the invariance of transformation is utilized via a rotation-guided refinement module.

We evaluate our method by adapting a pre-trained estimator to several dual-camera pairs placed in arbitrary poses \cite{H:ohkawa2023assemblyhands}.
Our evaluation encompasses both in-dataset and cross-dataset scenarios.
Experimental results reveal that our technique not only realizes notable improvements across all pairs but also surpasses state-of-the-art adaptation methods.
The primary contributions of this paper are summarized as:
\begin{itemize}
\item We propose a novel unsupervised single-to-dual-view adaptation (\proposed) solution for egocentric 3D hand pose estimation. Our method can adapt a traditional single-view estimator for arbitrary dual views without requiring annotations or camera parameters.
\item We build a pseudo-label-based strategy for adaptation. It leverages cross-view consensus and invariance of transformation between both camera coordinate systems for reliable pseudo-labeling. This leads to two key modules: attention-based merging and rotation-guided refinement.
\item Evaluation results demonstrate the benefits of our approach for arbitrarily placed camera pairs. Our method achieves significant improvements for all pairs both under in-dataset and cross-dataset settings.
\end{itemize}
\section{Related Work}
\label{sec:related}


\subsection{Multi-view hand pose estimation}
Multi-view hand pose estimation accepts multi-view images as input and outputs a final 3D hand pose, which remains a relatively unexplored research area.
Chen \etal \cite{H:chen2021mvhm} design a graph U-Net to integrate 3D hand poses from multi-view images.
Han \etal \cite{H:han2022umetrack} propose a unified multi-view fusion architecture to predict 3D hand poses. 
Method by Khaleghi \etal \cite{H:khaleghi2022multi} utilizes temporal information for multi-view fusion, outputting temporal hand pose predictions. 

All these studies have two limitations:
1) they require costly multi-view images and annotations for training, 2) during testing, camera poses are assumed to be known and covered by training data, thereby limiting their applicability. 
In contrast, our method eliminates the need for multi-view annotations and is adaptable to arbitrary dual views.

\subsection{Adaptation in hand pose estimation}
Adaptation aims at tailoring a model for specific application scenarios \cite{D:ohkawa2022domain,D:liu2022jitter,D:liang2021source}.
Existing adaptation methods in hand pose estimation mainly focus on adaptation across different domains (datasets), \eg, entropy minimization \cite{D:peng2023source,D:jiang2021regressive}, consistency regularization \cite{D:chen2019unsupervised,D:ohkawa2022domain,D:lin2023cross}, and pseudo-labeling methods \cite{D:huang2023semi,D:raychaudhuri2023prior,H:yang2021semihand}.
Prior works only propose constraints in single-view settings for adaptation, \eg, bio-mechanical constraint \cite{D:lin2023cross,H:yang2021semihand}, and are limited to single-view inference.
Unlike these methods, this paper proposes new stereo constraints from dual views for adaptation and supports dual-view inference, which extends the application scenarios.

\begin{figure}[t]
\begin{center}
    \includegraphics[width=.98\linewidth]{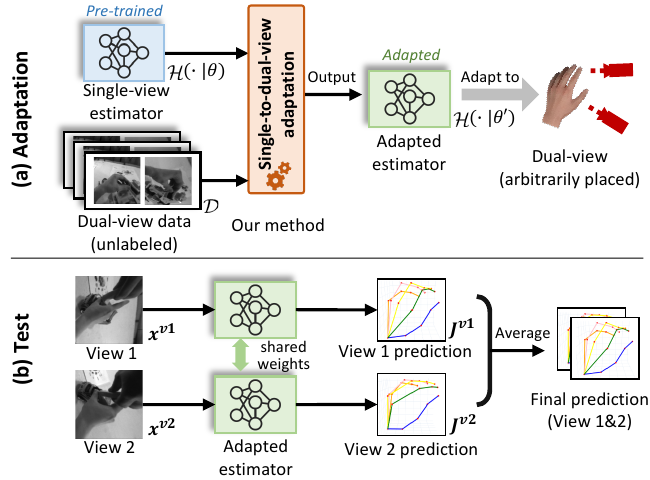}
\end{center}
\vspace{-5mm}
\caption{Problem setting of single-to-dual-view adaptation for hand pose estimation. (a) The input and output of adaptation. (b) The dual-view testing scheme after adaptation.}
\label{fig:task-def}
\vspace{-6mm}
\end{figure}

\section{Problem Setting}
\label{sec:taskdef}

\cref{fig:task-def} illustrates the task setting of single-to-dual-view adaptation for hand pose estimation.
We denote unlabeled dual-view data as  $\mathcal{D}=\{\mathbf{x}^{v1}_i, \mathbf{x}^{v2}_i|_{i=1}^{N}\}$, where $\mathbf{x}^{v1}_i$ and $\mathbf{x}^{v2}_i$ denote the \textit{i}-th image from view1 and view2, respectively, $N$ is the number of image pairs.
The dual-view data $\mathcal{D}$ contains no ground-truth hand poses or camera parameters.

As shown in \cref{fig:task-def} (a), suppose we have a baseline hand pose estimator $\mathcal{H}(\cdot|\theta)$ with parameters $\theta$ pre-trained from common single-view data.
Leveraging $\mathcal{D}$ can enhance its performance, as $\mathcal{D}$ provides additional information from a dual-view setup.
Our objective is to adapt this pre-trained estimator, $\mathcal{H}(\cdot|\theta)$, to an arbitrary yet fixed dual-view setting (with unknown camera poses) without needing ground-truths or camera parameters.
By inputting $\mathcal{H}(\cdot|\theta)$ and $\mathcal{D}$ into our method, it outputs an adapted estimator $\mathcal{H}(\cdot|\theta')$ with parameters $\theta'$ tailored for the dual-view scenario.

Upon adapting the estimator, its inference mechanism is correspondingly tailored for dual-view scenarios (\cref{fig:teaser} (b)).
During testing, the adapted estimator $\mathcal{H}(\cdot|\theta')$ processes a dual-view input pair $(\mathbf{x}^{v1},\mathbf{x}^{v2})$ and produces two predictions $(\mathbf{J}^{v1},\mathbf{J}^{v2})$, where each $\mathbf{J}^{v}\in\mathbb{R}^{21\times3}$ represents the 21 3D joints of the hand.
These predictions denote the 3D hand joints for each view and can be combined together to generate a final output, \eg, through a simple average.

\begin{figure}[t]
\begin{center}
\includegraphics[width=0.85\linewidth]{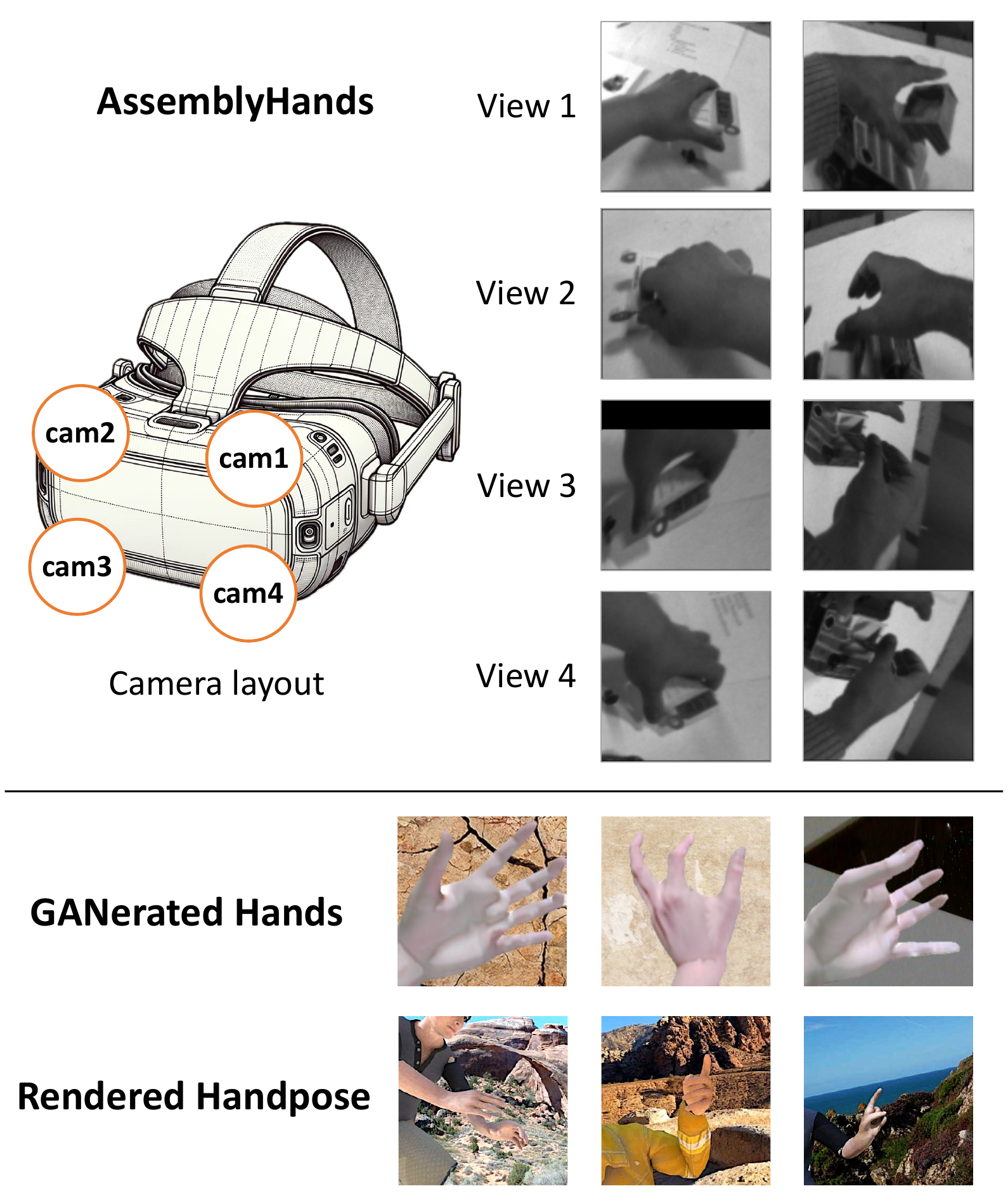}
\end{center}
\vspace{-5mm}
\caption{Top: Headset and its camera layout to collect multi-view data, and samples from the four views. Bottom: Samples of synthetic data. Image samples are from AssemblyHands \cite{H:ohkawa2023assemblyhands}, GANerated Hands \cite{H:GANeratedHands_CVPR2018}, and Rendered Handpose \cite{H:zb2017hand}, respectively.}
\label{fig:samples}
\vspace{-6mm}
\end{figure}

\noindent\textbf{Camera layout for a multi-view headset.}
\cref{fig:samples} illustrates an example of headset-mounted camera setups for multi-view egocentric data capture, with four cameras at each corner for different views. 
The top-right of \cref{fig:samples} displays images from these cameras. 
Six distinct dual-view pairs can be created from these four views. 
As a supplement, the bottom of \cref{fig:samples} shows the synthetic training data, highlighting variations in style and lighting. 
Such data helps to explore the performance of our method under cross-dataset or simulate-to-real settings.
See \cref{sec:dataset} for dataset details.

\section{Proposed Method}
\label{sec:methodology}
\begin{figure*}[t]
	\begin{center}
		\includegraphics[width=0.85\linewidth]{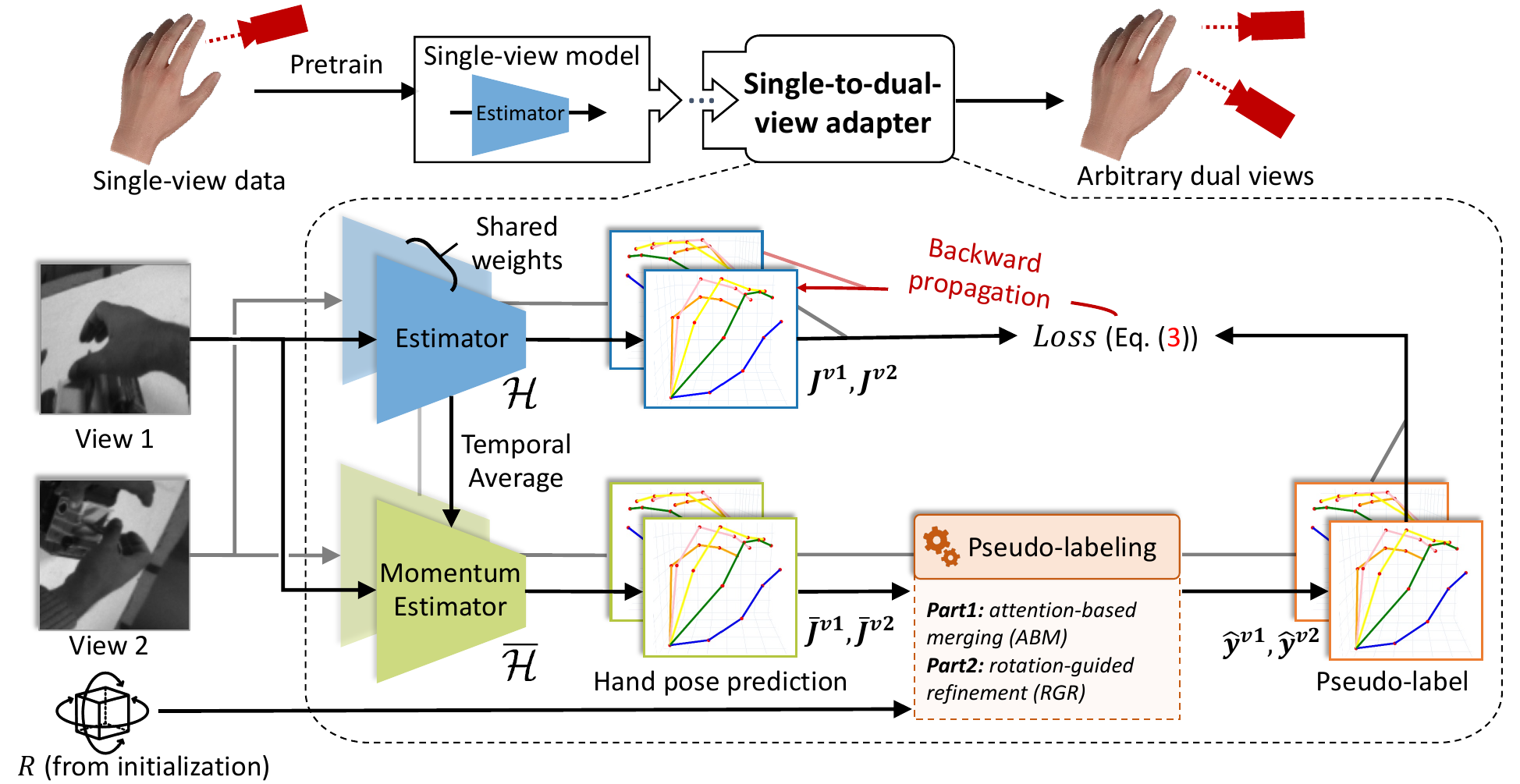}
	\end{center}
	\vspace{-5mm}
	\caption{Overview of the proposed \proposed, image pairs captured from arbitrarily placed dual cameras are input for adaptation. The architecture of \proposed~is illustrated in the dark dashed box, which contains a dynamically updated estimator and a momentum estimator. The momentum estimator's predictions are used to generate pseudo-labels, which are then processed by our pseudo-labeling module (\cref{sec:merge,sec:refine}). Using the pseudo-labels, a loss function is computed to update the estimator. The rotation matrix $R$ from the initialization step (\cref{sec:initialization}) is required for the pseudo-labeling.
	}
	\vspace{-5mm}
	\label{fig:overview}
\end{figure*}

We propose a novel unsupervised single-to-dual-view adaptation framework (\proposed). Before adaptation, an initialization step is performed to initialize the rotation matrix between both views (\cref{sec:initialization}).
The rotation matrix is essential to establish the transformation between two camera coordinate systems.
The architecture overview, as illustrated in \cref{fig:overview}, comprises two branches, an estimator $\mathcal{H}$ and a its momentum version $\overline{\mathcal{H}}$. 
The adaptation process is designed from two stereo constraints, pair-wise cross-view consensus and invariant rotation transformation between both camera coordinate systems.
This leads to two key pseudo-labeling modules: attention-based merging and rotation-guided refinement (\cref{sec:merge,sec:refine}).
Notably, these two modules function in a complementary manner, depending on the prediction accuracy, ensuring reliable pseudo-labeling.

\subsection{Initialization} \label{sec:initialization}
The initialization step aims to estimate a relatively accurate rotation matrix $R$, since $R$ is necessary to link the two camera coordinate systems \cite{N:zhang2023structural,N:liu2023uvagaze}.
It should be noted that translation vector between the two cameras is not necessary, as the predicted hand poses are usually aligned by the wrist during testing \cite{H:ohkawa2023assemblyhands,H:zhou2020monocular}.
Assuming that the initial pre-trained estimator is sufficiently accurate to generate reasonable predictions, we estimate the $R$ using the predictions of unlabeled dual-view data $\mathcal{D}=\{\mathbf{x}^{v1}_i, \mathbf{x}^{v2}_i|_{i=1}^{N}\}$.
Given $\mathcal{D}$, the estimator $\mathcal{H}$ can output $N$ pairs of predictions $\{\mathbf{J}_i^{v1}, \mathbf{J}_i^{v2}|_{i=1}^{N}\}$, where $\mathbf{J}_i^{v}\in\mathbb{R}^{21\times3}$ (21 is the number of 3D joints).
Then, the rotation matrix $R$ is estimated by:
\begin{equation} \label{eq:initialization}
	R^{(0)} = \frac{1}{N}\sum_{i=1}^{N} rot(\mathbf{J}_i^{v1}, \mathbf{J}_i^{v2}),
\end{equation}

\noindent where the superscript of $R$ denotes the iteration number, $(0)$ indicates that it is before the first iteration. 
The $rot$ function \cite{N:kabsch1978discussion} generates a $3\times3$ rotation matrix from two $21\times3$ joint predictions.
Note that the average in \cref{eq:initialization} is not element-wise, but an average of rotation matrices.

\subsection{Single-to-dual-view adaptation}
With the initialized $R$, the adaptation process begins.
The \proposed~framework comprises two branches, an estimator $\mathcal{H}(\cdot|\theta)$ with dynamically updating parameters $\theta$, and its momentum version $\overline{\mathcal{H}}(\cdot|\overline{\theta})$, which updates its parameters $\overline{\theta}$ using temporal moving average.
Temporal moving average has been proved by many works \cite{D:dubourvieux2021unsupervised,D:liu2021generalizing,N:he2020momentum,D:liu2024pnp} that can help to stabilize the training process.
The $\overline{\theta}$ is updated as:
\begin{equation} \label{eq:tem}
	\overline{\theta}^{(T)} = \eta_{\theta} \overline{\theta}^{(T-1)} + (1-\eta_{\theta})\theta,
\end{equation}

\noindent where $\overline{\theta}^{(T-1)}$ indicates the temporal averaged parameters in the previous iteration $T-1$, and $\eta_{\theta}$ represents the ensembling momentum, which is set as 0.99 \cite{D:liu2021generalizing,D:liu2024pnp}.

As shown in \cref{fig:overview}, during the single-to-dual-view adaptation, the role of the momentum model $\overline{\mathcal{H}}$ is to generate pseudo-labels, which are then utilized to supervise the model $\mathcal{H}$.
The pseudo-labeling module (\cref{sec:merge,sec:refine}) outputs pseudo-labels $\hat{\mathbf{y}}^{v1},\hat{\mathbf{y}}^{v2}$ based on the predictions $\mathbf{\overline{J}}^{v1}, \mathbf{\overline{J}}^{v2}$ from $\overline{\mathcal{H}}$.
These pseudo-labels are then used to supervise the predictions $\mathbf{J}^{v1}, \mathbf{J}^{v2}$ of $\mathcal{H}$.
The loss function is computed as:
\begin{equation} \label{eq:loss}
	\mathcal{L} = \| \mathbf{J}^{v1} - \hat{\mathbf{y}}^{v1} \|_2 + \| \mathbf{J}^{v2} - \hat{\mathbf{y}}^{v2} \|_2.
\end{equation}

\noindent The estimator follows the implementation of DetNet \cite{H:zhou2020monocular}, where $\mathcal{H}$ directly outputs heatmaps, and $\mathbf{J}$ is calculated from the heatmaps. Therefore, the loss function is actually computed from corresponding heatmaps, here we write these 3D-joint variables only for better understanding.

\begin{figure}[t]
	\begin{center}
		\includegraphics[width=0.9\linewidth]{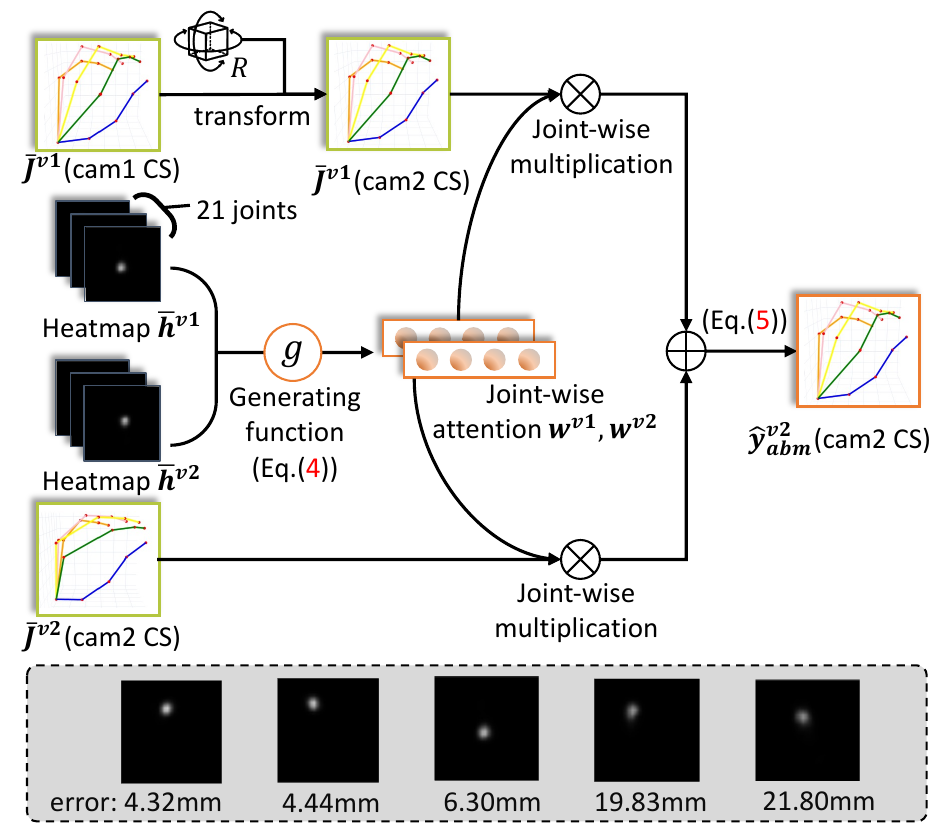}
	\end{center}
	\vspace{-5mm}
	\caption{Top: illustration of the first part of pseudo-labeling: attention-based merging module.
		The generating process of $\hat{y}_{abm}^{v2}$ in view2 is shown as an example, the process of view1 is the same.
		Bottom: visualizations of heatmaps with different accuracy.
	}
	\label{fig:merging}
	\vspace{-6mm}
\end{figure}

\subsection{Pseudo-labeling: attention-based merging} \label{sec:merge}
The attention-based merging (ABM) module, which constitutes the first part of pseudo-labeling, is derived from cross-view consensus.
Cross-view consensus refers to the concept of achieving agreement or consistency between different views of the same data \cite{N:wu2023semi}.
Theoretically, when transformed into the same coordinate system, the two predictions $\mathbf{\overline{J}}^{v1}$ and $\mathbf{\overline{J}}^{v2}$ from different views should be identical, \ie, $R\mathbf{\overline{J}}^{v1}=\mathbf{\overline{J}}^{v2}$, with $\mathbf{\overline{J}}^{v1}$ $\mathbf{\overline{J}}^{v2}$ being aligned with wrist joint.

This stereo constraint is the foundation for this module to generate accurate pseudo-labels.
Prior works utilize a simple average \cite{D:ohkawa2022domain,H:liu2021semi} (\eg, $(R\mathbf{\overline{J}}^{v1}+\mathbf{\overline{J}}^{v2})/2$) or sample-wise confidence \cite{D:ohkawa2022domain,D:cai2020generalizing} to improve the quality of pseudo-labels. 
However, these approaches overlook the varying confidence in joints that is caused by differences in image capture across views.
For instance, a joint that is occluded in one view but fully visible in another could lead to reliable predictions being hindered by unreliable ones.

To address this, we propose joint-wise attention $\mathbf{w}\in\mathbb{R}^{21\times1}$ to represent each joint's prediction confidence.
It is derived from the 2D heatmap $\mathbf{\overline{h}} \in \mathbb{R}^{21\times32\times32}$ output from $\overline{\mathcal{H}}$.
The $\mathbf{\overline{h}}$ indicates the probability of each joint's presence at every pixel in the 2D image space.
This approach is from an observation (bottom of \cref{fig:merging}): as the error of prediction increases, the intensity of the heatmap's hotspot decreases, \ie, darker indicates low accuracy.
Inspired by this, we propose an attention-generating function:
\begin{equation} \label{eq:attention}
	\mathbf{w}_j^v = \frac{\beta^{max(\mathbf{\overline{h}}_j^v)}}{\sum_{v\in\{v1,v2\}}\beta^{max(\mathbf{\overline{h}}_j^v)}},
\end{equation}

\noindent where the subscript $j$ indicate the index of joint, \ie, $j=1,2,...,21$. 
We introduce a hyper-parameter $\beta$ here to adjust softness.
Please refer to \cref{sec:hyper} for parameter choosing.

The workflow of this attention-based merging module is illustrated at the top of \cref{fig:merging}.
First, we transform both predictions into the same camera coordinate system.
Then, a joint-wise multiplication is performed for each of them using the attention $\mathbf{w}^{v1},\mathbf{w}^{v2}$.
Finally, the pseudo-label $\mathbf{\hat{y}}_{abm}$ is calculated through a summation operation.
In summary:
\begin{equation} \label{eq:yabm}
	\begin{aligned}
		\mathbf{\hat{y}}_{abm}^{v1}&=\mathbf{w}^{v1}\overline{\mathbf{J}}^{v1}+\mathbf{w}^{v2}R^T\overline{\mathbf{J}}^{v2}, \\
		\mathbf{\hat{y}}_{abm}^{v2}&=\mathbf{w}^{v1}R\overline{\mathbf{J}}^{v1}+\mathbf{w}^{v2}\overline{\mathbf{J}}^{v2}.
	\end{aligned}
\end{equation}

\subsection{Pseudo-labeling: rotation-guided refinement} \label{sec:refine}
The rotation-guided refinement (RGR) module is based on another stereo constraint: invariance of rotation transformation between both views. 
This implies that the estimated rotation matrix should remain unchanged since the cameras are fixed, \ie, theoretically $rot(\mathbf{\overline{J}}^{v1},\mathbf{\overline{J}}^{v2})=C$.
In light of this, this module aims to refine the predictions $\mathbf{\overline{J}}^{v1},\mathbf{\overline{J}}^{v2}$ such that $rot(\mathbf{\overline{J}}^{v1},\mathbf{\overline{J}}^{v2})$ becomes invariant across all input data.
The refinement result becomes the pseudo-label $\mathbf{\hat{y}}_{rgr}$.

\begin{figure}[t]
	\begin{center}
		\includegraphics[width=.95\linewidth]{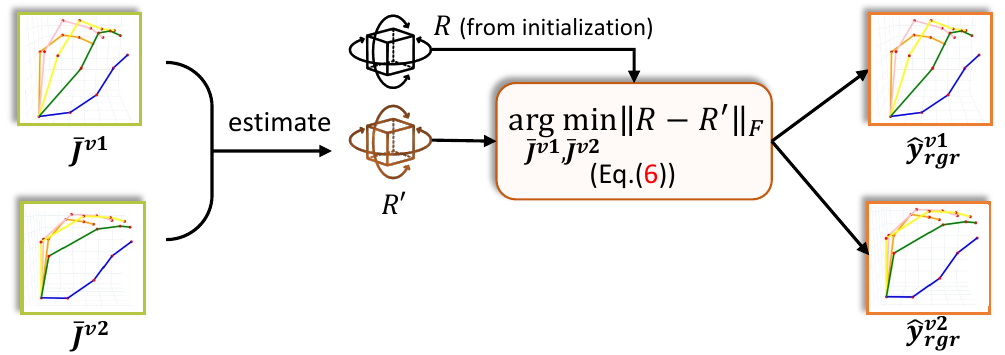}
	\end{center}
	\vspace{-5mm}
	\caption{Illustration of the second part of pseudo-labeling: rotation-guided refinement module.
	}
	\label{fig:refinement}
	\vspace{-6mm}
\end{figure}

The workflow is shown in \cref{fig:refinement}.
Given a pair of predictions, our method estimates a new rotation matrix $R'=rot(\mathbf{\overline{J}}^{v1},\mathbf{\overline{J}}^{v2})$.
Subsequently, the more accurately estimated $R$ from the initialization step is set as the target for refinement, aiming to minimize $\|R - R'\|_F$.
The refinement process can be expressed as:
\begin{equation} \label{eq:yrgr}
	\mathbf{\hat{y}}_{rgr}^{v1}, \mathbf{\hat{y}}_{rgr}^{v2}=\underset{{\mathbf{\overline{J}}^{v1},\mathbf{\overline{J}}^{v2}}}{\arg\min}\|R-rot(\mathbf{\overline{J}}^{v1},\mathbf{\overline{J}}^{v2})\|_F.
\end{equation}

\noindent In detail, we employ BFGS \cite{N:broyden1970convergence} algorithm for minimizing. 

\noindent\textbf{Final pseudo-label.}
Finally, the pseudo-label is calculated from a weighted average of $\mathbf{\hat{y}}_{abm}$ and $\mathbf{\hat{y}}_{rgr}$:
\begin{equation} \label{eq:pseudo}
	\mathbf{\hat{y}}=\alpha\mathbf{\hat{y}}_{abm} + (1-\alpha)\mathbf{\hat{y}}_{rgr}.
\end{equation}

\noindent Here, we introduce another pre-fixed hyper-parameter $\alpha$, which adjusts the the weight of the two parts of the pseudo-label. Empirically, we set $\alpha=0.7$ (see \cref{sec:hyper}).

\noindent\textbf{Complement between two pseudo-labels.}
When the predictions $\mathbf{\overline{J}}^{v1}$ and $\mathbf{\overline{J}}^{v2}$ are accurate, $R'$ closely approximates $R$, making $\mathbf{\hat{y}}_{rgr}$ redundant. 
In such cases, $\mathbf{\hat{y}}_{abm}$ is beneficial as it merges the two accurate predictions. 
Conversely, if $\mathbf{\overline{J}}^{v1}$ and $\mathbf{\overline{J}}^{v2}$ are unreliable, $\mathbf{\hat{y}}{abm}$ will consequently also lack accuracy. 
In such instances, $\mathbf{\hat{y}}{rgr}$ steps in as a complementary tool for refining pseudo-labels.
By minimizing $\|R-R'\|_F$, it optimizes the predictions towards an alignment with real-world condition.

\noindent\textbf{Update rotation matrix R.}
Clearly, the rotation matrix $R$ from the initialization step plays an important role in bridging the two camera coordinate systems.
As can be expected, its accuracy significantly affects the final performance.
To enhance its accuracy with each iteration, we also employ a temporal moving average for its updates.
Given an input batch containing $B$ image pairs, the $R$ is updated as:
\begin{equation} \label{eq:update-r}
	R^{(T)} = \eta_R R^{(T-1)} + (1-\eta_R)\cdot\frac{1}{B}\sum_{i=1}^{B} rot(\mathbf{\overline{J}}_i^{v1}, \mathbf{\overline{J}}_i^{v2}),
\end{equation}

\noindent where the ensembling momentum $\eta_R$ is set as 0.999 \cite{N:he2020momentum}.
This way, the rotation matrix $R$ can be updated slowly and provide a more accurate rotation function with iteration.

\begin{table*}[t]
\caption{Adaptation results for all dual-camera pairs. The camera pairs are from $\mathcal{D}_{ah}$ dataset \cite{H:ohkawa2023assemblyhands}, which is divided into two parts according to the collecting headset. In ``In-dataset" and ``Cross-dataset" settings, the baseline model is pre-trained on $\mathcal{D}_{ah}$ and $\mathcal{D}_{syn}$ \cite{H:zb2017hand,H:GANeratedHands_CVPR2018}, respectively. The adaptation yields a uniquely adapted model for each pair, and ``Overall" averages the results across all 6 pairs.}
\vspace{-6mm} \label{tab:merged-pair}
\begin{center}
\small
\renewcommand\arraystretch{.95}
\setlength{\tabcolsep}{1mm}
\resizebox{.98\linewidth}{!}{
\begin{tabular}{cl|cc|cc||cc|cc}
    \toprule[1.2pt]
    \multirow{3}{*}{Camera pair} & \multirow{3}{*}{Method} & \multicolumn{4}{c}{\underline{In-dataset ($\mathcal{D}_{ah} \rightarrow \mathcal{D}_{ah}$)}} & \multicolumn{4}{c}{\underline{Cross-dataset ($\mathcal{D}_{syn} \rightarrow \mathcal{D}_{ah}$)}} \\
    & & \multicolumn{2}{c}{$\mathcal{D}_{ah}-Headset1$} & \multicolumn{2}{c}{$\mathcal{D}_{ah}-Headset2$} & \multicolumn{2}{c}{$\mathcal{D}_{ah}-Headset1$} & \multicolumn{2}{c}{$\mathcal{D}_{ah}-Headset2$} \\
    & & \mono & \avg & \mono & \avg & \mono & \avg & \mono & \avg \\
    \hline
    \specialrule{0em}{1pt}{1pt}
    \multirow{2}{*}{$cam~1,2$} 
    & Baseline        & 43.00 & 39.20 & 54.71 & 52.38 & 67.93 & 60.48 & 70.32 & 62.26 \\
    & \textbf{\proposed} & \textbf{31.01} \upscore{27.9} & \textbf{31.36} \upscore{20.0} & \textbf{45.52} \upscore{16.8} & \textbf{45.14} \upscore{13.8} & \textbf{63.46} \upscore{6.6} & \textbf{59.32} \upscore{1.9} & \textbf{70.09} \upscore{0.3} & \textbf{60.97} \upscore{2.1}\\
    \cline{1-10} \specialrule{0em}{1pt}{1pt}

    \multirow{2}{*}{$cam~1,3$}
    & Baseline        & 25.00 & 23.29 & 22.59 & 21.08 & 57.79 & 51.42 & 64.00 & 60.25\\
    & \textbf{\proposed} & \textbf{19.73} \upscore{21.1} & \textbf{19.92} \upscore{14.5} & \textbf{17.90} \upscore{20.8} & \textbf{17.68} \upscore{16.1} & \textbf{50.84} \upscore{12.0} & \textbf{47.55} \upscore{7.5} & \textbf{61.81} \upscore{3.4} & \textbf{58.34} \upscore{3.2}\\
    \cline{1-10} \specialrule{0em}{1pt}{1pt}

    \multirow{2}{*}{$cam~1,4$}
    & Baseline        & 24.90 & 22.70 & 16.73 & 14.91 & 52.71 & 46.55 & 54.32 & 50.57\\
    & \textbf{\proposed} & \textbf{20.88} \upscore{16.1} & \textbf{20.87} \upscore{8.1} & \textbf{14.64} \upscore{12.5} & \textbf{14.29} \upscore{4.2} & \textbf{46.05} \upscore{12.6} & \textbf{42.50} \upscore{8.7} & \textbf{46.59} \upscore{14.2} & \textbf{45.66} \upscore{9.7} \\
    \cline{1-10} \specialrule{0em}{1pt}{1pt}

    \multirow{2}{*}{$cam~2,3$}
    & Baseline        & 17.96 & 15.23 & 17.10 & 15.08 & 53.36 & 48.42 & 52.84 & 48.84\\
    & \textbf{\proposed} & \textbf{14.97} \upscore{16.6} & \textbf{14.44} \upscore{5.2} & \textbf{14.42} \upscore{15.7} & \textbf{14.20} \upscore{5.8} & \textbf{40.26} \upscore{24.6} & \textbf{39.32} \upscore{18.8} & \textbf{43.61} \upscore{17.5} & \textbf{42.88} \upscore{12.2} \\
    \cline{1-10} \specialrule{0em}{1pt}{1pt}
    
    \multirow{2}{*}{$cam~2,4$}
    & Baseline        & 22.09 & 19.84 & 23.24 & 20.96 & 59.44 & 54.32 & 61.13 & 57.41 \\
    & \textbf{\proposed} & \textbf{17.98} \upscore{18.6} & \textbf{17.75} \upscore{10.5} & \textbf{18.31} \upscore{21.2} & \textbf{18.41} \upscore{12.2} & \textbf{50.59} \upscore{14.9} & \textbf{49.41} \upscore{9.0} & \textbf{52.45} \upscore{14.2} & \textbf{51.48} \upscore{10.3}\\
    \cline{1-10} \specialrule{0em}{1pt}{1pt}

    \multirow{2}{*}{$cam~3,4$}
    & Baseline        & 16.83 & 15.77 & 19.93 & 18.08 & 45.82 & 42.34 & 49.84 & 48.99\\
    & \textbf{\proposed} & \textbf{16.36} \upscore{2.8} & \textbf{15.55} \upscore{1.4} & \textbf{19.25} \upscore{3.4} & \textbf{17.80} \upscore{1.5} & \textbf{39.46} \upscore{13.9} & \textbf{37.43} \upscore{11.6} & \textbf{44.04} \upscore{11.6} & \textbf{42.88} \upscore{12.5}\\
    \cline{1-10} \specialrule{0em}{1pt}{1pt}

    \multirow{2}{*}{Overall}
    & Baseline        & 24.96 & 22.67 & 25.72 & 23.75 & 56.18 & 50.59 & 58.74 & 54.72\\
    & \textbf{\proposed} & \textbf{20.16} \upscore{19.2} & \textbf{19.98} \upscore{11.9} & \textbf{21.67} \upscore{15.7} & \textbf{21.25} \upscore{10.5} & \textbf{48.44} \upscore{13.8} & \textbf{45.92} \upscore{9.2} & \textbf{53.11} \upscore{9.6} & \textbf{50.37} \upscore{7.9}\\
    
    \bottomrule[1.2pt]
\end{tabular}}
\end{center}
\vspace{-7mm}
\end{table*}

\section{Experiment}
\label{sec:experiment}

\subsection{Dataset} \label{sec:dataset}
We employ \textbf{AssemblyHands} \cite{H:ohkawa2023assemblyhands} ($\mathcal{D}_{ah}$) as the evaluation set, as it is the newest large-scale benchmark dataset with high-quality multi-view 3D hand pose annotations. 
As for the training set,
we set two adaptation scenarios: 1) in-dataset setting where the training set is drawn from the same dataset $\mathcal{D}_{ah}$ and 2) cross-dataset setting where we use synthetic dataset ($\mathcal{D}_{syn}$) as the training set, consisting of \textbf{Rendered Handpose} \cite{H:zb2017hand} and \textbf{GANerated Hands} \cite{H:GANeratedHands_CVPR2018}.
The details of the datasets are as below:
\begin{itemize}
\item\textit{AssemblyHands} \cite{H:ohkawa2023assemblyhands} is a large-scale benchmark dataset featuring accurate 3D hand pose annotations. Collected using two AR headsets, it comprises images captured from four synchronized egocentric cameras. The dataset includes 412K training samples and 62K testing samples.
\item\textit{GANerated Hands} \cite{H:GANeratedHands_CVPR2018} 
includes over 330K color images of hands. The images are synthetically generated and then fed to a GAN \cite{N:goodfellow2014generative} to make the features closer to real hands. 
\item\textit{Rendered Handpose} \cite{H:zb2017hand} contains about 44K samples. The images are rendered with freely available characters.
\end{itemize}
In detail, AssemblyHands is collected by two VR headsets, as shown in \cref{fig:samples}, each headset has four egocentric cameras at four corners. 
Following the collecting devices, we separate $\mathcal{D}_{ah}$ into two parts, each part is collected using one headset, namely $\mathcal{D}_{ah}-Headset1/2$.

\subsection{Experimental setup} 
\textbf{Evaluation metric.} We compare the predictions from our model with the ground-truth labels in root-relative coordinates, and use the common mean per joint position error (MPJPE) in millimeters as the evaluation metric. 
However, since our focus is on the single-to-dual-view adaptation task, the adapted estimator is expected to perform in dual-view settings.
This implies that traditional MPJPE computed from single-view (monocular MPJPE, \mono) cannot be sufficient. 
As a result, we propose a new dual-view MPJPE metric \avg~in addition to monocular MPJPE. The metrics are defined as follows:
\begin{itemize}
\item \mono: the traditional monocular MPJPE, which collects all the single-view errors from both views and calculates their average.
\item \avg: the proposed metric under dual views. To calculate it, first the predictions from both views are averaged using a rotation matrix $R$. Then, we calculate the MPJPE of the averaged predictions as the \avg. Usually, the $R$ is from the initialization step of our method.
\end{itemize}
\noindent \textbf{Implementation detail.} We employ PyTorch for implementation. All experiments run on a single NVIDIA A100 GPU.
DetNet from \cite{H:zhou2020monocular} is adopted as the backbone of our hand pose estimator.
Adam optimizer is employed with a learning rate of $1\times10^{-3}$ to
pre-train the 3D hand pose estimation network. 
For the single-to-dual-view adaptation, we use the Adam optimizer with a learning rate of $5\times10^{-4}$.

\subsection{Adaptation results for all camera pairs}
In this section, we use our method to adapt the same pre-trained single-view hand pose estimator to all dual-view pairs from the evaluation set $\mathcal{D}_{ah}$ independently, yielding one adapted model for each pair. 
Experiments are conducted under both in-dataset and cross-dataset settings.
Under the in-dataset setting (\cref{tab:merged-pair}, $\mathcal{D}_{ah} \rightarrow \mathcal{D}_{ah} $), the baseline model is pre-trained on $\mathcal{D}_{ah}$, while under the cross-dataset setting ($\mathcal{D}_{syn} \rightarrow \mathcal{D}_{ah} $), the baseline model is pre-trained on $\mathcal{D}_{syn}$ before being adapted to the camera pairs from $\mathcal{D}_{ah}$.

As shown in \cref{tab:merged-pair}, compared with the pre-trained model (Baseline), our \proposed~offers significant accuracy gains under both settings among all camera pairs. 
This indicates that our method can adapt well to arbitrary dual views regardless of the camera positions or pre-training datasets. 

Quantitative results demonstrate that the \proposed~offers substantial improvements.
On average, the improvement in both monocular (\mono) and dual-view (\avg) metrics exceeds 10\%, with the maximum improvement exceeding 20\%.
Interestingly, we can see that the improvement for $cam~1,2$ under the cross-dataset setting is relatively small. 
This indicates that low initial accuracy limits the performance.

\subsection{Comparison under cross-dataset settings}
Our method is compared with state-of-the-art adaptation techniques in cross-dataset settings. 
Considering the prevalence and significance of cross-dataset scenarios in real-world applications, this experiment evaluates the capability of \proposed~in comparison to leading domain adaptation methods.
Specifically, adaptation methods included in the comparison are: SFDAHPE \cite{D:peng2023source}, RegDA \cite{D:jiang2021regressive}, DAGEN \cite{D:guo2020domain}, and ADDA \cite{D:tzeng2017adversarial}.

For fairness, we do not include existing multi-view methods \cite{H:chen2021mvhm,H:khaleghi2022multi,H:han2022umetrack} in this comparison.
This is because these methods require 1) multi-view labels and 2) camera parameters, whereas our approach is unsupervised and does not require such parameters.
In contrast, all the comparison methods are unsupervised, leading to a fair comparison.


In detail, SFDAHPE \cite{D:peng2023source}, RegDA \cite{D:jiang2021regressive} are developed for pose estimation.
DAGEN \cite{D:guo2020domain} and ADDA \cite{D:tzeng2017adversarial} are originally proposed for gaze estimation and classification, respectively. 
We include these two methods here to show potential of these state-of-the-art methods in enhancing the cross-dataset performance of hand pose estimation. 
To make a fair comparison, their original networks are replaced with the same DetNet \cite{H:zhou2020monocular} as our baseline.

Quantitative results of different methods are shown in \cref{tab:sota}. 
Our method not only significantly outperforms the state-of-the-art methods, but also shows an advantage of being source-free.
The superior performance verifies the effectiveness of the proposed \proposed~for single-to-dual-view adaptation under cross-dataset settings.
For reference, we also provide the result of fine-tuning as the upper bound of this cross-dataset task.

\begin{table}
\caption{Comparison with state-of-the-art adaptation methods under cross-dataset settings. 
``SF" indicates if the method is source-free (requiring no data from source dataset, $\mathcal{D}_{syn}$). 
* denotes that labels from the target dataset ($\mathcal{D}_{ah}$) are needed.
}
\label{tab:sota}
\vspace{-6mm}
\setlength{\tabcolsep}{1mm}{
\begin{center}
\renewcommand\arraystretch{1}
\resizebox{\linewidth}{!}{
\begin{tabular}{lccc|cc}
    \toprule[1.2pt]
    \multirow{2}{*}{$\mathcal{D}_{syn} \rightarrow \mathcal{D}_{ah}$} & & \multicolumn{2}{c}{\underline{$\mathcal{D}_{ah}-Headset1$}} & \multicolumn{2}{c}{\underline{$\mathcal{D}_{ah}-Headset2$}}  \\
    & SF & \mono & \avg & \mono & \avg \\
    \hline  \specialrule{0em}{1pt}{1pt}
    Source Only &  & 56.18 & 50.59 & 58.74 & 54.72 \\
    Fine-tune* &  & 45.03 & 38.11 & 47.75 & 42.19 \\
    \hline  \specialrule{0em}{1pt}{1pt}
    ADDA \cite{D:tzeng2017adversarial} & \ding{55} & 56.90 & 48.48 & 57.87 & 51.39 \\
    DAGEN \cite{D:guo2020domain} & \ding{55} & 55.37 & 49.72 & 57.62 & 53.17 \\
    RegDA \cite{D:jiang2021regressive} & \ding{55} & 51.41 & 47.85 & 54.75 & 51.50 \\
    SFDAHPE \cite{D:peng2023source} & \ding{51} & 54.06 & 49.11 & 57.22 & 53.39 \\
    \textbf{\proposed~(Ours)} & \ding{51} & \textbf{48.44} & \textbf{45.92} & \textbf{53.11} & \textbf{50.37} \\
    \bottomrule[1.2pt]
\end{tabular}
}
\end{center}}
\vspace{-5mm}
\end{table}

\begin{table}
\caption{Ablation study of our method on $\mathcal{D}_{ah} \rightarrow \mathcal{D}_{ah}$ task. ABM and RGR stand for the two pseudo-labeling modules, respectively.
}
\label{tab:ablation}
\vspace{-6mm}
\setlength{\tabcolsep}{2mm}{
\begin{center}
\renewcommand\arraystretch{1}
	\resizebox{.95\linewidth}{!}{
    \begin{tabular}{cccc|cc}
        \toprule[1.2pt]
        \multirow{2}{*}{ABM} & \multirow{2}{*}{RGR} & \multicolumn{2}{c}{\underline{$\mathcal{D}_{ah}-Headset1$}} & \multicolumn{2}{c}{\underline{$\mathcal{D}_{ah}-Headset2$}}  \\
        & & \mono & \avg & \mono & \avg \\
        \hline  \specialrule{0em}{1pt}{1pt}
        \textcolor{red}{\ding{55}} & \textcolor{red}{\ding{55}} & 24.96 & 22.67 & 25.72 & 23.75 \\
        \hline  \specialrule{0em}{1pt}{1pt}
        \textcolor{darkgreen}{\ding{51}} & \textcolor{red}{\ding{55}} & 20.81 & 20.54 & 22.24 & 21.71 \\
        \textcolor{red}{\ding{55}} & \textcolor{darkgreen}{\ding{51}} & 21.89 & 21.33 & 23.54 & 22.75 \\
        \textcolor{darkgreen}{\ding{51}} & \textcolor{darkgreen}{\ding{51}} & \textbf{20.16} & \textbf{19.98} & \textbf{21.67} & \textbf{21.25} \\
        \bottomrule[1.2pt]
    \end{tabular}
    			}
\end{center}}
\vspace{-7mm}
\end{table}

\subsection{Ablation study}
We conducted ablation experiments to analyze the contribution of each component in our model. 
The following experiments are evaluated based on the $\mathcal{D}_{ah} \rightarrow \mathcal{D}_{ah}$ task for clearer observation. 
The components are shown below:
\begin{itemize}
    \setlength
    \item ABM: Attention-based merging module, which generates pseudo-labels based on the cross-view consensus.
    \item RGR: Rotation-guided refinement module, which generates pseudo-labels based on the invariance of rotation transformation between both camera coordinate systems.
\end{itemize}
\cref{tab:ablation} shows the hand pose estimation errors under different combinations. 
We observe that both ABM and RGR can significantly improve the hand pose estimation performance over the pre-trained baseline (first row). 
Our final version achieves the best results for all metrics, confirming the optimality of our method.

\subsection{Number of input image pairs}
\begin{figure}
	\centering
	\includegraphics[width=\linewidth]{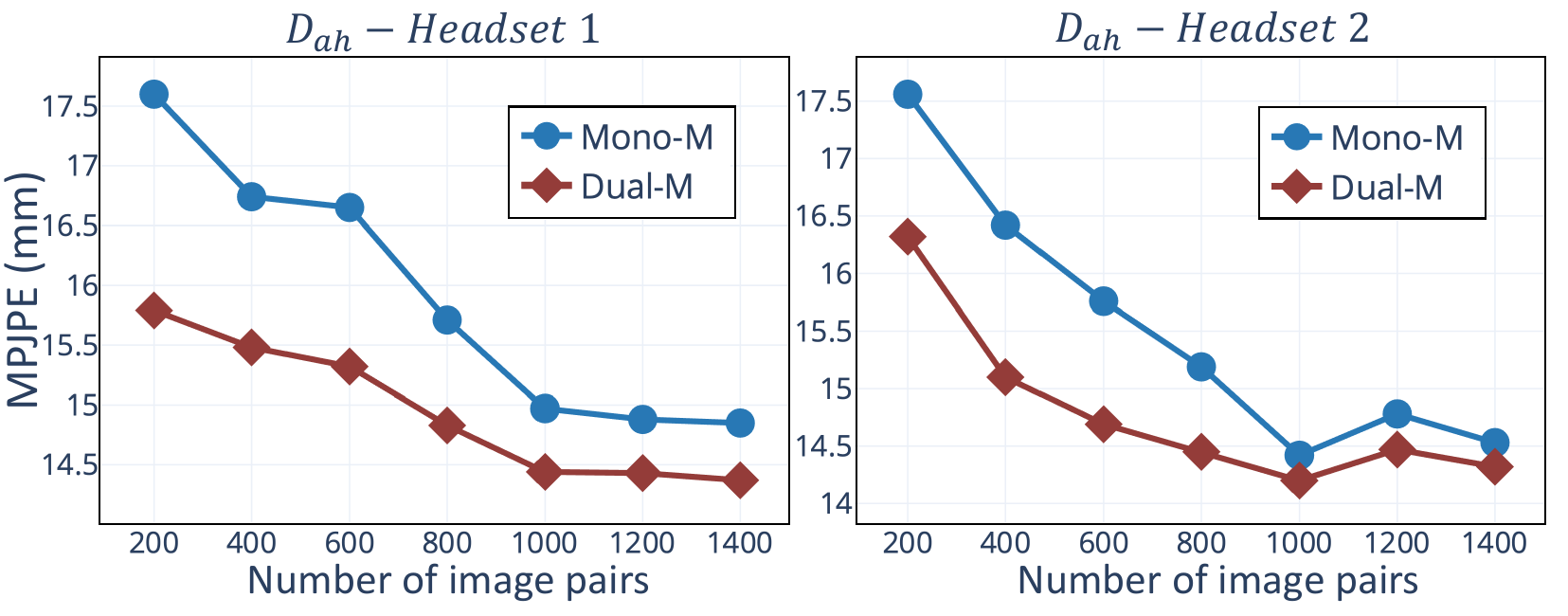}
	\vspace{-7mm}
	\caption{Evaluation of \proposed's performance with gradually increasing the number of input image pairs.} 
	\label{fig:pairnum}
	\vspace{-2mm}
\end{figure}

To find the optimal number of input image pairs (\ie, $N$) for our method, we evaluate the \proposed's performance under different numbers of input image pairs. 
Specifically, the experiments are conducted on the $cam~2,3$ pair in the $\mathcal{D}_{ah} \rightarrow \mathcal{D}_{ah}$ task. 
The results are illustrated in \cref{fig:pairnum}.
It indicates that the performance of \proposed~constantly improves as the number of input image pairs increase. 
Our method's performance converges when $N\geq1000$. 
Consequently, we choose $N=1000$ for our \proposed.

\subsection{Complement between two pseudo-labels} \label{sec:complement}
\begin{figure}
	\centering
	\includegraphics[width=.95\linewidth]{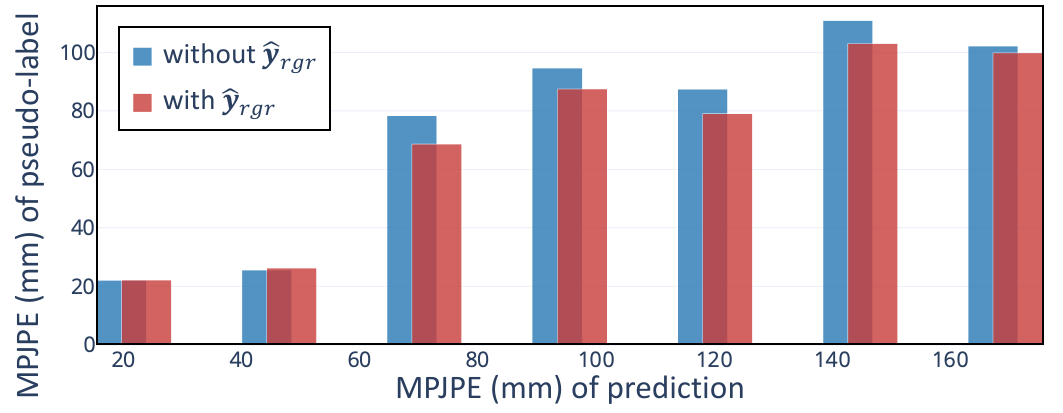}
	\vspace{-3mm}
	\caption{The error of pseudo-labels with and without $\hat{\mathbf{y}}_{rgr}$, in relation to the error of prediction $\mathbf{\overline{J}}$, where $\mathbf{\overline{J}}$ is what we use to compute the pseudo-labels. MPJPE in millimeter is the metric.} 
	\label{fig:complement}
	\vspace{-2mm}
\end{figure}

\cref{fig:complement} demonstrates the complementary nature, as stated in \cref{sec:refine}, of the pseudo-labeling.
Using camera pair $cam1,2-Headset1$ under in-dataset setting, we analyze the error of pseudo-labels with and without the refinement term $\hat{\mathbf{y}}_{rgr}$, in relation to the prediction error of $\mathbf{\overline{J}}$. 
Note that $\mathbf{\overline{J}}$ is the prediction used to compute these pseudo-labels.
For better observation, the predictions are first divided into seven equal intervals according to their errors, $[9.4,34.0), [34.0, 58.6),...,[157.1,181.7]$. 
Then, the average pseudo-label error for each interval is calculated.

In \cref{fig:complement}, the bars are placed in the middle of each interval, with y-axis representing the pseudo-label errors.
The result indicates that the $\hat{\mathbf{y}}_{rgr}$ term is redundant for accurate predictions ($<60mm$) but significantly reduces pseudo-label errors for larger prediction errors ($\geq60mm$).
This finding supports the statement in \cref{sec:refine} about the importance of $\hat{\mathbf{y}}_{rgr}$ for complementing pseudo-labels in cases of inaccurate predictions.

\begin{table}
\caption{Performance of our method with different hyper-parameters $\alpha$ (\cref{eq:pseudo}) and $\beta$ (\cref{eq:attention}).
}
\label{tab:hyper}
\vspace{-5mm}
\setlength{\tabcolsep}{2mm}{
\begin{center}
\renewcommand\arraystretch{.95}
\resizebox{.95\linewidth}{!}{
\begin{tabular}{lcc|cc}
    \toprule[1.2pt]
    \multirow{2}{*}{$\mathcal{D}_{ah} \rightarrow \mathcal{D}_{ah}$} & \multicolumn{2}{c}{\underline{$\mathcal{D}_{ah}-Headset1$}} & \multicolumn{2}{c}{\underline{$\mathcal{D}_{ah}-Headset2$}}  \\
    & \mono & \avg & \mono & \avg \\
    \hline  \specialrule{0em}{1pt}{1pt}
    $\alpha=0.3$ & 20.83 & 20.54 & 22.39 & 21.82 \\
    $\alpha=0.5$ & 20.34 & 20.17 & 21.90 & 21.49 \\
    $\alpha=0.7$ & \textbf{20.16} & \textbf{19.98} & \textbf{21.67} & \textbf{21.25} \\
    $\alpha=0.9$ & 20.46 & 20.18 & 22.17 & 21.53 \\
    \hline  \specialrule{0em}{1pt}{1pt}
    $\beta=1$ & 21.02 & 20.94 & 22.77 & 22.56 \\
    $\beta=e$ & 20.66 & 20.47 & 22.50 & 22.10 \\
    $\beta=\infty$ & \textbf{20.16} & \textbf{19.98} & \textbf{21.67} & \textbf{21.25} \\
    \bottomrule[1.2pt]
\end{tabular}
            }
\end{center}}
\vspace{-8mm}
\end{table}

\subsection{Hyper-parameters} \label{sec:hyper}
We evaluate how the \proposed's performance varies with the change of weight parameter $\alpha$ (\cref{eq:pseudo}). 
$\alpha$ controls the weights for averaging the two parts of pseudo-labeling. 
We test four values of $\alpha$, 0.3, 0.5, 0.7, and 0.9. 
The results are shown in the row 1-4 of \cref{tab:hyper}, where our method achieves the best performance when $\alpha=0.7$.

We also test the performance with varying $\beta$ (\cref{eq:attention}). 
$\beta$ is the parameter in the attention-generating function, which generates the joint-wise attention in attention-based merging module. 
In fact, the $\beta$ acts the role of $e$ in the softmax function. 
When $\beta=1,e,\infty$, the merging becomes a simple average (where attention becomes invalidated), softmax function, and maximum function, respectively. 
We can see that the \proposed~achieves the best performance when $\beta=\infty$.
This suggests that selecting the prediction with higher confidence as a pseudo-label in a joint-wise manner is the most effective strategy.
Consequently, we set $\alpha=0.7$ and $\beta=\infty$ for all the experiments.

\subsection{Qualitative result}
\begin{figure}[t]
	\centering
	\includegraphics[width=.95\linewidth]{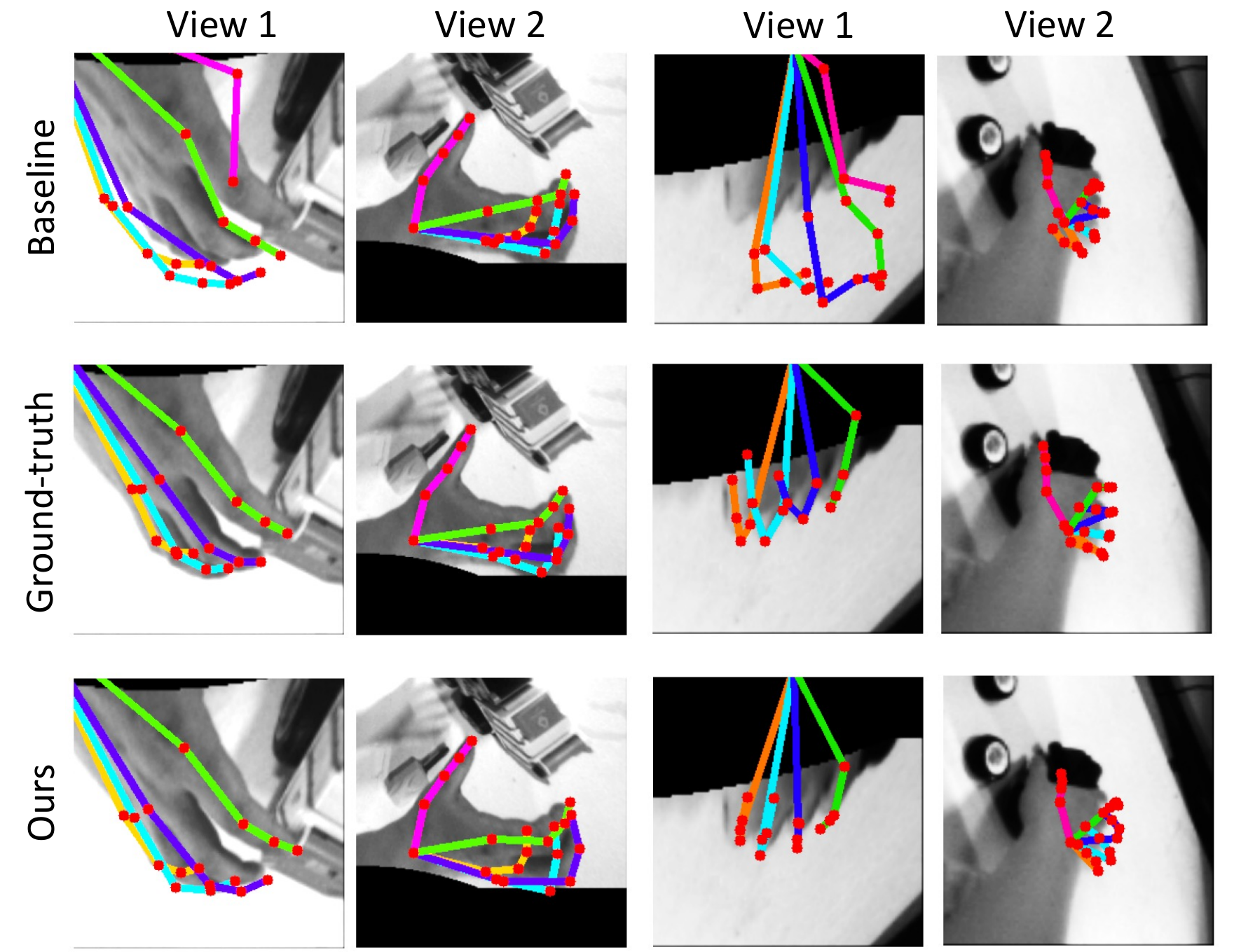}
	\vspace{-3mm}
	\caption{Visual examples of estimated 3D hand poses under both views. The joints are portrayed by projecting the final 3D predictions to the image plane.}
	\label{fig:visualpose}
	\vspace{-6mm}
\end{figure}

To understand how our method improves the performance of hand pose estimation under dual-view settings,  we visually present typical cases by portraying 3D hand joints onto the input image pairs.
In detail, the 3D joints are projected to the image plane, with the visualization depicted in \cref{fig:visualpose}.

Notably, when confronted with extreme view angles (see the left pair), the predictions of baseline model tend to be unreliable. 
Conversely, our technique gives a prediction much closer to the actual hand shape after adaptation.
In the 3rd column, even when the hand is partially out of field-of-view, leading to a truncated hand, our \proposed~continues to deliver trustworthy predictions.
These results indicate that \proposed~can utilize additional information from dual views to provide significant improvements even under extreme challenging cases.

\section{Conclusion}
In this paper, we present a novel single-to-dual-view adaptation framework (\proposed), designed to adapt a single-view hand pose estimator to dual-view settings. 
The \proposed~is unsupervised, eliminating the need for multi-view labels. 
Our method also requires no camera parameters, enabling compatibility with arbitrary dual views. 
Two stereo constraints are employed as two pseudo-labeling modules in an complementary manner.
Our method achieves significant performance gains across all dual-view pairs under both in-dataset and cross-dataset settings.

{
    \small
    \bibliographystyle{ieeenat_fullname}
    \bibliography{main}
}


\end{document}